\documentclass[10pt, conference]{IEEEtran}
 \usepackage[utf8]{inputenc}
\usepackage{siunitx}

\usepackage[hyphens]{url}
\usepackage[breaklinks]{hyperref}
\hypersetup{hidelinks}
\usepackage{amsmath, 
	amsfonts, 
	amssymb, 
	bm,
	mathtools,
	amsthm,
	siunitx}
\usepackage{graphicx}
\usepackage{
		xcolor
	}
\usepackage{booktabs}
\usepackage{multirow}
\usepackage{tikz,
	pgfplots,
	pgfplotstable,
	eso-pic}
\usetikzlibrary{patterns,
	plotmarks}
\pgfplotsset{compat=1.12}

\usepackage[english,
	onelanguage,
	ruled,
	vlined,
	longend,
	titlenotnumbered]{algorithm2e}

\usepackage[nospace]{cite}

\usepackage{blindtext,
	ifpdf}
\usepackage{setspace}
\usepackage{paralist}
\usepackage{filecontents}

\newcommand{\bfa}{\mathbf{a}}

\newcommand{\bfb}{\mathbf{b}}
\newcommand{\hbfb}{\hat{\mathbf{b}}}

\newcommand{\bfe}{\mathbf{e}}

\newcommand{\bfF}{\mathbf{F}}
\newcommand{\bfG}{\mathbf{G}}

\newcommand{\bfg}{\mathbf{g}}
\newcommand{\bfh}{\mathbf{h}}
\newcommand{\bfH}{\mathbf{H}}
\newcommand{\bfI}{\mathbf{I}}

\newcommand{\bfK}{\mathbf{K}}

\newcommand{\bfN}{\mathbf{N}}
\newcommand{\calN}{\mathcal{N}}

\newcommand{\bfp}{\mathbf{p}}

\newcommand{\bfP}{\mathbf{P}}

\newcommand{\bfQ}{\mathbf{Q}}

\newcommand{\bfR}{\mathbf{R}}

\newcommand{\bbR}{\mathbb{R}}

\newcommand{\hbfu}{\hat{\mathbf{u}}}
\newcommand{\bfv}{\mathbf{v}}

\newcommand{\bfw}{\mathbf{w}}

\newcommand{\bfx}{\mathbf{x}}

\newcommand{\hbfx}{\hat{\mathbf{x}}}

\newcommand{\bfy}{\mathbf{y}}

\newcommand{\hbfy}{\hat{\mathbf{y}}}
\newcommand{\bfz}{\mathbf{z}}

\newcommand{\hz}{\hat{z}}

\newcommand{\hbfz}{\hat{\bfz}}

\newcommand{\bochi}{\protect\raisebox{1pt}{$\boldsymbol{\chi}$}}
\newcommand{\hbochi}{\hat{\bochi}}

\newcommand{\boomega}{\boldsymbol{\omega}}

\newcommand{\boxi}{\boldsymbol{\xi}}

\newcommand{\bfzero}{\mathbf{0}}

\DeclareMathOperator{\diag}{diag}

\newtheoremstyle{mystyle}
{}
{}
{\itshape}
{}
{\bfseries}
{.}
{ }
{}

\usetikzlibrary{shapes,arrows,shadows}
\pgfdeclarelayer{background}
\pgfdeclarelayer{foreground}
\pgfsetlayers{background,main,foreground}

\tikzstyle{wa} = [draw, text width=10em, fill=red!20, 
minimum height=4em, rounded corners, drop shadow, text centered]
\tikzstyle{de} = [draw, text width=10em, fill=blue!20, 
minimum height=4em, rounded corners, text centered,drop shadow]

\usepackage{afterpage}
\usepackage{times}
\usepackage{filecontents}

\usepackage[top=54pt, bottom=54pt, left=54pt, right=54pt]{geometry}

\usepgfplotslibrary{external} 
\begin{document}

	\title{\vspace*{0.5cm}RINS-W: Robust Inertial Navigation System on Wheels}
	
	\author{\IEEEauthorblockN{
			Martin \textsc{Brossard}\IEEEauthorrefmark{1}
			, Axel  \textsc{Barrau}\IEEEauthorrefmark{2}
			and Silv\`ere \textsc{Bonnabel}\IEEEauthorrefmark{1} 
		}
		\IEEEauthorblockA{\IEEEauthorrefmark{1}MINES ParisTech, PSL Research University, Centre for Robotics, 60 Boulevard Saint-Michel, 75006 Paris, France}
		\IEEEauthorblockA{\IEEEauthorrefmark{2}Safran Tech, Groupe Safran, Rue des Jeunes Bois-Ch\^ateaufort, 78772, Magny Les Hameaux Cedex, France}	
	}
	
	\maketitle	
	
	\begin{abstract}
		This paper proposes a real-time approach for long-term inertial navigation based \emph{only} on an Inertial Measurement Unit (IMU) for self-localizing wheeled robots. The approach builds upon two components: 1) a robust detector  that uses  recurrent deep neural networks to dynamically detect a variety of situations of interest, such as zero velocity or no lateral slip; and 2) a state-of-the-art  Kalman filter which  incorporates this knowledge as pseudo-measurements  for localization. Evaluations on a publicly available car dataset demonstrates that the proposed scheme may achieve a final precision of 20 m for a 21 km long trajectory of a vehicle driving for over an hour, equipped with an IMU of moderate precision (the gyro drift rate is 10 deg/h). To our knowledge, this is the first paper which combines sophisticated deep learning techniques with state-of-the-art filtering methods for pure inertial navigation on wheeled vehicles and as such opens up  for novel data-driven inertial navigation techniques. Moreover, albeit taylored for IMU-only based localization, our method may be used as a component for self-localization of wheeled robots equipped with a more complete sensor suite.  
	\end{abstract}
	
	
	\section{Introduction}\label{sec:int}
	
	Inertial navigation, or ``inertial odometry", uses an Inertial Measurement Unit (IMU)  consisting of accelerometers and gyrometers to  calculate in real time by dead reckoning the position, the orientation, and the 3D velocity vector of a moving object without the need for external references or extra  sensors. Albeit  a mature field, the development of  low-cost and   small size inertial sensors  over the past two decades has attracted much interest for  robotics and autonomous systems applications, see e.g.  \cite{oxts, collinInertial2018, kokUsing2017}.

	High precision Inertial Navigation Systems (INS) achieve very small localization errors but are costly and rely on time-consuming initialization procedures \cite{safranInertial2018}.  In contrast, MEMS-based IMU suffer from large errors such as scale factor, axis misalignment, thermo-mechanical white  noise and random walk  noise, resulting in rapid localization drift. This prompts the need for fusion with complementary sensors such as  GPS, cameras, or LiDAR, see e.g.,  \cite{cadenaPresent2016,sunRobust2018, deschaudIMLSSLAM2018}.
	
\begin{figure*}
		\centering
		\begin{tikzpicture}[]
		\begin{axis}[height=11.5cm,
		width=18.cm,
		ylabel=$y$ (\SI{}{km}),
		xlabel=$x$ (\SI{}{km}),
		xmax=2.65,
		xmin=-0.45,
		ymax=1.05,
		ymin=-1.3,
		ylabel style={xshift=0.0cm,yshift=-0.0cm},
		xlabel style={xshift=0cm,yshift=0cm},
		legend columns=2, legend pos= north east, 
		label style={font=\small}, 
		ticks=both,
		legend entries={ground truth, IMU, odometry, odometry+FoG, \textbf{proposed RINS-W}},
		legend style={font=\small}, 
		yticklabel style={
			/pgf/number format/fixed,
			/pgf/number format/precision=1,
			font=\small
		},
		xticklabel style={font=\small,
			/pgf/number format/fixed,
			/pgf/number format/precision=1,}
		]
		
		\addplot[draw = black, 
		very thick] table[x=x,y=y] {figures/16gt.txt};
		\addplot[draw = cyan, dashed, thick] table[x=x,y=y] {figures/16imu.txt};
		\addplot[draw = red, dashed, thick] table[x=x,y=y] {figures/16odo.txt};
		\addplot[draw = blue, dashed, thick] table[x=x,y=y] {figures/16odofog.txt};
		\addplot[draw = green, dashed, thick] table[x=x,y=y] {figures/16hat.txt};
		\path [draw, ->] (-0.1, -0.3)  node [below] {start} to[bend  right=-40,looseness=1.3, ->] (-0.01, -0.01);
		
		\node at (0.49,-1.04) {\Large \color{red}\textbf{*}};
		\node at (0.33,-0.02) {\Large \color{blue}\textbf{*}};
		\node at (0.4016,0.1819) {\Large \color{green}\textbf{*}};
		
		\path [draw, ->] (0.4, 0.6)  node [left] {ground truth end} to[bend  right=-80,looseness=1., ->] (0.41, 0.16);
		\path [draw, ->] (0.5, 0.8)  node [right] {first turn} to[bend  right=30,looseness=0.5, ->] (-0.1, 0.85);
		\end{axis}	
		\end{tikzpicture}
		\vspace*{-0.7cm}\caption{Trajectory ground truth and estimates obtained by various methods: integration of IMU signals; odometry based on a differential wheel encoder system; odometry combined with an highly accurate and expensive Fiber optics Gyro (FoG) that provides orientation estimates; and the proposed RINS-W approach which considers only the IMU sensor embarked in the vehicle, and which outperforms the other schemes. The final distance error for this long-term sequence \texttt{urban16} (73 minutes) of the car dataset \cite{jeongComplex2018}  is \SI{20}{m} for the RINS-W solution. The deep learning based detector (see Section \ref{sec:detector}) has of course \emph{not} been trained or cross-validated on this sequence. \label{fig:results}}
		\vspace*{-0.6cm}
	\end{figure*}
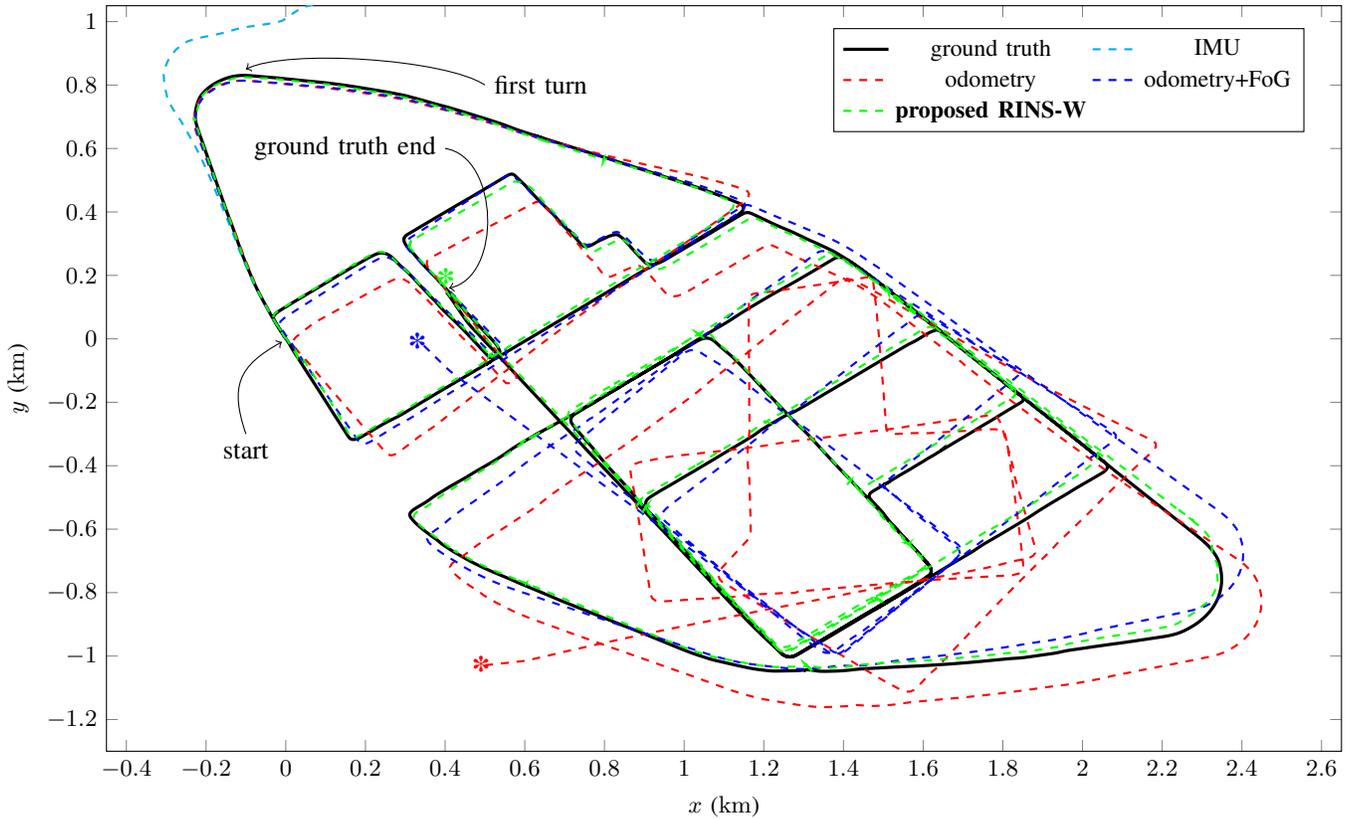

	In this paper, we provide a method for  long-term localization for wheeled robots only using an IMU. Our contributions, and the paper's organization, are as follows:
	\begin{itemize}
		\item we introduce  specific motion profiles frequently encountered by a wheeled vehicle  which bring information about the motion  when correctly identified, along with  their mathematical description in Section \ref{sec:sub};
		\item we design an algorithm which automatically detects in real time   if any of those assumptions about the motion holds in Section \ref{sec:detector}, based only on the IMU measurements. The detector builds upon  recurrent deep neural networks \cite{goodfellowDeep2016} and is trained using IMU  and ground truth data;
		\item we implement a state-of-the-art invariant extended Kalman filter \cite{barrauInvariant2017,barrauInvariant2018} that exploits the detector's outputs as pseudo-measurements to combine them with the IMU  outputs  in a statistical way, in Section \ref{sec:filter}. It yields  accurate estimates of   pose, velocity and sensor biases, along with associated uncertainty (covariance);
		\item we demonstrate the performances of the approach on a publicly available car dataset \cite{jeongComplex2018} in Section \ref{sec:results}. Our approach solely based on the IMU produces accurate estimates  with a final distance w.r.t. ground truth of \SI{20}{m} on the 73 minutes test sequence \texttt{urban16}, see Figure \ref{fig:results}. In particular, our approach  outperforms differential wheel odometry, as well as wheel odometry aided by an expensive fiber optics  gyro  whose  drift  is  200  times  smaller  than the one of the IMU we use;
		\item this evidences that accurately detecting some specific patterns in the IMU data and incorporating this information in a filter may prove very fruitful for localization;
		\item the method is not restricted to IMU only based navigation.  It is possible to feed the Kalman filter  with other sensors' measurements. Besides, once trained the detector may be used as a building block in any localization algorithm.  
	\end{itemize}

	\subsection{Related Works}
	Odometry and localization based on inertial sensors, cameras, and/or LIDARs  have made tremendous progresses over the last decade, enabling robust real-time localization systems, see e.g., \cite{cadenaPresent2016, sunRobust2018, deschaudIMLSSLAM2018}. As concerns   back-end techniques,  although optimization-based methods tend to  presently prevail, filtering based methods building upon the  Invariant Extended Kalman Filter (IEKF)  \cite{barrauInvariant2017, barrauInvariant2018} are currently  gaining  interest, owing to its theoretical guarantees in terms of convergence and consistency. The IEKF \cite{barrauInvariant2017} has already given rise to a commercial product in the field of high precision navigation, see  \cite{barrauAligment2016,barrauInvariant2018}. It has also been successfully applied to  visual inertial odometry, in  \cite{wuinvariantEKF2017,heoConsistent2018,brossardUnscented2018} and bipedal robotics \cite{hartleyContactAided2018}. 
	
	Taking into account vehicle constraints and odometer measurements are known to increase the robustness of   visual-inertial systems \cite{wuVINS2017, zhengOdometry2018}. Although quite successful,  systems using vision continuously process a large amount of data which is computationally demanding and energy consuming. Moreover, an autonomous vehicle should run in parallel its own robust IMU-based localization algorithm to   perform maneuvers such as  emergency stops  if LiDAR and camera are unavailable due to lack of texture or information,  and more generally   failures \cite{oxts}.
	
	Inertial navigation systems  have long leveraged  pseudo-measurements from IMU signals, e.g. the widespread Zero velocity UPdaTe (ZUPT) \cite{ramanandanInertial2012, skogZeroVelocity2010}. Upon detection of a zero-velocity event, such pseudo-measurement can be fused with the dead reckoning motion model in an extended Kalman filter \cite{dissanayakeAiding2001, solinInertial2018} or in a factor graph, see \cite{atchuthanOdometry2018}.
	
	Very recently,  great efforts have been devoted to the use  of deep learning and more generally  machine learning frameworks for pedestrian inertial navigation \cite{yanRIDI2018, wagstaffLSTMBased2018, cortesDeep2018, chenIONet2018, chenOxIOD2018}. In \cite{yanRIDI2018} velocity is estimated using support vector regression whereas \cite{wagstaffLSTMBased2018,cortesDeep2018, chenIONet2018} use recurrent neural networks respectively for ZUPT detection, speed estimation, and end-to-end inertial navigation. Those methods are promising but difficult to transfer to a wheeled robots since they generally only consider horizontal planar motion, and must infer velocity directly from a small sequence of IMU measurements, whereas we can afford to use larger sequences. Finally, in \cite{brossardlearning2019}, we used Gaussian Processes to learn and correct wheel encoders errors  to improve wheel encoder based dead reckoning in 2D.

	\section{Inertial Navigation System \& Sensor Model}\label{sec:model}
	
	Denoting the IMU orientation  by $\bfR_n \in SO(3)$, i.e. the rotation matrix that maps the body frame to the world frame $\mathrm{\textsc{w}}$, its velocity in $\mathrm{\textsc{w}}$ by $\bfv_n^{\mathrm{\textsc{w}}} \in \bbR^3$ and its position in $\mathrm{\textsc{w}}$  by $\bfp_n^{\mathrm{\textsc{w}}} \in \bbR^3$, the dynamics are as follows
	\begin{align}
	\bfR_{n+1} &= \bfR_n \exp_{SO(3)}\left(\boomega_n dt\right), \label{eq:prop1}\\
	\bfv_{n+1}^{\mathrm{\textsc{w}}} &= \bfv_n^{\mathrm{\textsc{w}}} + \left(\bfR_n\bfa_n + \bfg \right)dt, \label{eq:prop2}\\
	\bfp_{n+1}^{\mathrm{\textsc{w}}} &= \bfp_n^{\mathrm{\textsc{w}}} +  \bfv_n^{\mathrm{\textsc{w}}} dt,\label{eq:prop3}
	\end{align}
	between two discrete times, with sampling time $dt$, and where   $\left(\bfR_0, \bfv_0^{\mathrm{\textsc{w}}}, \bfp_0^{\mathrm{\textsc{w}}}\right)$ is the initial configuration. The true angular velocity $\boomega_n \in \bbR^3$ and the true specific acceleration $\bfa_n \in \bbR^3$ are the  inputs of the system \eqref{eq:prop1}-\eqref{eq:prop3}. In our application scenarios, the effects of earth rotation    and  Coriolis acceleration are ignored, and earth is considered flat.
	
	\subsection{Inertial Measurement Unit (IMU) Model}
	
	The IMU provides noisy and biased measurements of $\boomega_n \in \bbR^3$ and $\bfa_n \in \bbR^3$ as follows
	\begin{align}
	\boomega_n^{\mathrm{\textsc{imu}}} &= \boomega_n + \bfb_{n}^{\boomega} + \bfw_{n}^{\boomega}, \label{eq:gyro}\\
	\bfa^{\mathrm{\textsc{imu}}}_n &= \bfa_n + \bfb_{n}^{\bfa} + \bfw_{n}^{\bfa}, \label{eq:acc}
	\end{align}
	where $\bfb_{n}^{\boomega}$, $\bfb_{n}^{\bfa}$ are quasi-constant biases and $\bfw_{n}^{\boomega}$, $\bfw_{n}^{\bfa}$ are zero-mean Gaussian noises. The biases follow a random-walk
	\begin{align}
	\bfb_{n+1}^{\boomega} &= \bfb_{n}^{\boomega} + \bfw^{\bfb_{\omega}}_{n}, \\
	\bfb_{n+1}^{\bfa} &= \bfb_{n}^{\bfa} + \bfw^{\bfb_{\bfa}}_{n}, \label{eq:propend}
	\end{align}
	where $\bfw_{n}^{\bfb_{\boomega}}$, $\bfw^{\bfb_{\bfa}}_{n}$ are zero-mean Gaussian noises.

	All sources of error - particularly biases - are yet undesirable since a simple implementation of \eqref{eq:prop1}-\eqref{eq:prop3} leads to  a triple integration of raw  data, which is much more harmful that the unique integration of differential wheel speeds.  Even a small error  may thus cause  the position estimate to be way off the true position,  within seconds.
	
	\section{Specific Motion Profiles for Wheeled Systems}\label{sec:sub}
	
	We describe in this section characteristic motions  frequently encountered by a wheeled robot,   that provide useful complementary information to the IMU when correctly detected.
	\subsection{Considered Motion Profiles}
	We consider 4 distinct specific motion profiles, whose validity is encoded in the following binary vector:
	\begin{align}
	\bfz_n = \left(z^{\mathrm{\textsc{vel}}}_n,~z^{\mathrm{\textsc{ang}}}_n,~ z^{\mathrm{\textsc{lat}}}_n,~ z^{\mathrm{\textsc{up}}}_n\right)\in\{0,1\}^4. \label{eq:z}
	\end{align}
	\begin{itemize}
		\item \textbf{Zero velocity}: the velocity of the platform is null. As a consequence so is  the linear acceleration, yielding:
		\begin{align}
		z^{\mathrm{\textsc{vel}}}_n = 1 \Rightarrow \left\lbrace \begin{matrix}
		\bfv_n = \bfzero \\
		\bfR_n \bfa_n + \bfg = \bfzero
		\end{matrix}\right. .\label{eq:z_vel}
		\end{align}
		Such profiles  frequently occur for  vehicles moving in  urban environments, and are leveraged in the well known Zero velocity UPdaTe (ZUPT), see e.g. \cite{ramanandanInertial2012, skogZeroVelocity2010}.
		\item \textbf{Zero angular velocity}: the heading is constant:
		\begin{align}
		z^{\mathrm{\textsc{ang}}}_n = 1 \Rightarrow \boomega_n = \bfzero. \label{eq:z_ang}
		\end{align}
		\item \textbf{Zero lateral velocity}\footnote{Without loss of generality, we assume that the body frame is aligned with the IMU frame.}: the lateral velocity is considered roughly null
		\begin{align}
		z^{\mathrm{\textsc{lat}}}_n = 1 \Rightarrow v_n^{\mathrm{\textsc{lat}}} \simeq 0, \label{eq:z_lat}
		\end{align}
		where we obtain the lateral velocity $v_n^{\mathrm{\textsc{lat}}}$  after expressing the velocity in the body frame $\mathrm{\textsc{b}}$ as
		\begin{align}
		\bfv_n^{\mathrm{\textsc{b}}} = \bfR_n^T \bfv_n^{\mathrm{\textsc{w}}} = \begin{bmatrix}
		v_n^{\mathrm{\textsc{for}}} \\
		v_n^{\mathrm{\textsc{lat}}} \\
		v_n^{\mathrm{\textsc{up}}}
		\end{bmatrix}.
		\end{align} 
		
		\item \textbf{Zero vertical velocity}: the vertical velocity is considered roughly null
		\begin{align}
		z^{\mathrm{\textsc{up}}}_n = 1 \Rightarrow v_n^{\mathrm{\textsc{up}}} \simeq 0. \label{eq:z_up}
		\end{align}
		
	\end{itemize}
	The two latter  are common assumptions for wheeled robots or cars moving forward on human made roads.  
	
	\subsection{Discussion on the Choice of Profiles}
	
\begin{figure}
		\center
		\begin{tikzpicture}[]
		\begin{axis}[height=5cm,
		width=9cm,
		xlabel=$t$ (s), ylabel= {\SI{e4}{m/s}, \SI{}{rad/s}},
		legend columns=3,  legend pos=
		north east, 
		label style={font=\small}, 
		ticks=both,
		xmin=0,
		xmax=8,
		legend entries={$\|\bfv^{\mathrm{\textsc{w}}}_n\|$, $\|\boomega_n\|$, $z^{\mathrm{\textsc{vel}}}_n$},
		legend style={font=\small}, 
		yticklabel style={
			/pgf/number format/fixed,
			/pgf/number format/precision=1,
			font=\small
		},
		xticklabel style={font=\small,
			/pgf/number format/fixed,
			/pgf/number format/precision=1,}
		]
		]
		\addplot[draw = blue, thick, smooth] table[x=t,y=v_norm] {figures/d8.txt};
		\addplot[draw = green, thick] table[x=t,y=ang_norm] {figures/d8.txt};
		\addplot[draw = magenta, thick ] table[x=t,y=zupt_gt] {figures/d8.txt};
		\end{axis}	
		\end{tikzpicture}
		\vspace*{-0.6cm}\caption{Real IMU data of a car stopping from sequence \texttt{urban06} of \cite{jeongComplex2018}. We see \eqref{eq:z_vel} holds and thus $	z^{\mathrm{\textsc{vel}}}_n=1$  at $t=5.8s$ while \eqref{eq:z_ang} \label{fig:z_ang} does not hold yet.} \vspace*{-0.3cm}
	\end{figure}
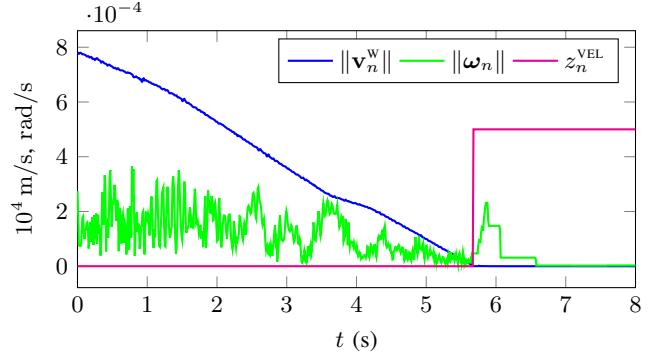

	The motion profiles were carefully chosen in order to match the specificity of wheeled robots  equipped with shock absorbers. Indeed, as illustrated  on Figure \ref{fig:z_ang}, when the wheels of a car actually stop, the car undergoes a rotational motion forward and then backward. As a result, null velocity \eqref{eq:z_vel} does not imply null angular velocity \eqref{eq:z_ang}. For the level of precision we pursue in the present paper, distinguishing between \eqref{eq:z_vel} and \eqref{eq:z_ang} is  pivotal since it allows us to: $i$) properly label motion profiles before training (see Section \ref{sec:train}, where we have different thresholds on position and on angular velocity); and: $ii$) improve detection accuracy since only one motion pattern can be identified as valid. 

	\eqref{eq:z_lat} and \eqref{eq:z_up} generally hold  for  robots moving indoors or cars   on roads.  Note that \eqref{eq:z_up} is expressed in the body frame, and thus generally holds for a car moving on a road even if the road is not level. As such  \eqref{eq:z_up} is more refined that just assuming the car is moving in a 2D horizontal plane.  And it is actually quite a  challenge for an IMU-based detector to identify when \eqref{eq:z_lat}  and   \eqref{eq:z_up} are \emph{not} valid.

	\subsection{Expected Impacts on  Robot Localization}
	
	The motion profiles fall into two categories in terms of the information they bring:
	\begin{enumerate}
		\item \textbf{Zero velocity constraints}:  the profiles \eqref{eq:z_vel}-\eqref{eq:z_ang}, when correctly detected  may allow to correct the IMU biases, the pitch and the roll.
		\item \textbf{Vehicle motion constraints}: the profiles  \eqref{eq:z_lat} and \eqref{eq:z_up} are useful constraints for the estimates accuracy over the long term. In particular, Section \ref{sec:resbonus} will experimentally demonstrate the benefits of accounting for   \eqref{eq:z_lat} and \eqref{eq:z_up}.
	\end{enumerate}

	\section{Proposed RINS-W Algorithm}\label{sec:system}
\begin{figure}
		\center
		\begin{tikzpicture}
    \node (wa) [wa]  {Invariant Extended Kalman Filter};
    \path (wa.west)+(-3,0.0) node[text width=1cm,align=center] (asr2) {$\boomega_n^{\mathrm{\textsc{imu}}}$  $\bfa_n^{\mathrm{\textsc{imu}}}$};
       
    \path (wa.east)+(1,0) node (vote)  {$\hbfx_{n+1}$};

    \path [draw, ->] (asr2.east) -- node [above] {} (wa.180);
    \path [draw, ->] (wa.east) -- node [above] {} (vote.west);
        
    \path (wa.north)+(-1.5,1.0) node[de] (de)  {Specific Motion Profile Detector};
    \path (wa.south) +(-1,-0.5) node (asrs) {RINS-W system};
  
  	 \path [draw, ->] (asr2.east) -- node [above] {}  (asr2.east)+(0.5,0) -- node [above] {}  ++(0.5,1.75) -- ++(0.4,0);
  	 
  	 \path [draw, ->] (de.east) -- ++(0.5,0) node [above] {$\hbfz_{n+1}$} -- ++(0,-1);
  
    \begin{pgfonlayer}{background}
        \path (wa.south)+(-4,-1) node (a) {};
        \path (de.north)+(3.75,0.5) node (c) {};
        \path[rounded corners, draw=black!50, dashed]
            (a) rectangle (c);           
    \end{pgfonlayer}
\end{tikzpicture}\vspace*{-0.5cm}
		\caption{Structure of the proposed RINS-W system for inertial navigation. The detector  identifies specific motion profiles \eqref{eq:z} from raw IMU signals only.  \label{fig:scheme}}\vspace*{-0.7cm}
	\end{figure}
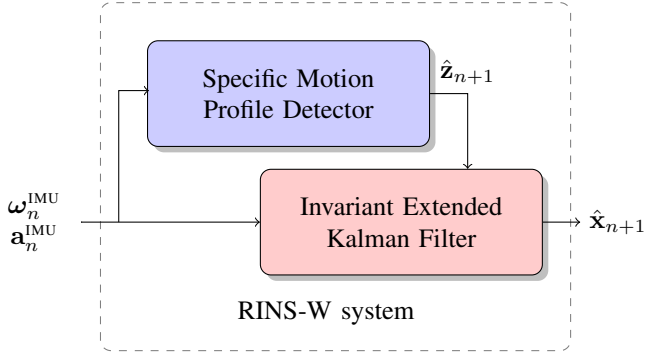

	This section describes our system for recovering trajectory and sensor bias estimates from an IMU. Figure \ref{fig:scheme} illustrates the approach which consists of two main blocks summarized as follow: 
	\begin{itemize}
		\item the detector estimates the binary vector $\bfz_n$ from raw IMU signals, and consists of recurrent neural networks;
		\item the filter integrates the IMU measurements in its dynamic model and exploits the detected motion profiles  as pseudo-measurements  to refine its estimates, as customary in inertial navigation, see e.g. \cite{ramanandanInertial2012}.
	\end{itemize}
	The detector does not use the filter's output and is based only on IMU measurements. Thus  the detector operates autonomously and is trained independently, see Section \ref{sec:train}. Note that our approach is different from more tightly coupled approaches such as \cite{haarnojaBackprop2016}. We now describe each block.
	
	\subsection{The Specific Motion Profile Detector}\label{sec:detector}
	
\begin{figure}
		\center
		\begin{tikzpicture}
    \node (wa) [draw, text width=4em,
    minimum height=2em, drop shadow, text centered,fill=green!20]  {LSTM};
    \path (wa.west)+(-1.5,0.0) node[text width=1cm,align=center] (asr2) {$\boomega_n^{\mathrm{\textsc{imu}}}$  $\bfa_n^{\mathrm{\textsc{imu}}}$};
    
    \path (wa.east)+(2,0.0) node (th) [draw, text width=4em,
    minimum height=2em, drop shadow, text centered,fill=yellow!20]  {Threshold};
    \path [draw, ->] (wa.east)+(0,-0.0) -- ++(1.16,-0.0);
    \path (th.east)+(1,0) node (vote)  {$\hbfz_{n+1}$};

    \path [draw, ->] (asr2.east) -- node [above] {} (wa.180);
    \path [draw, ->] (th.east) -- node [above] {} (vote.west);
        
    \path (wa.south) +(1.4,-0.6) node (asrs) {Specific Motion Profile Detector};

    \path [draw,densely dotted, ->] (wa.east)+(0,0.2) -- ++(0.2,0.2) to[bend  right=90,looseness=1.6, ->]  node [above] {$\bfh_{n+1}$} (wa.north);
    
    \begin{pgfonlayer}{background}
        \path (wa.south)+(-1.25,-1) node (a) {};
        \path (wa.north)+(4,1) node (c) {};
        \path[rounded corners, draw=black!50, dashed, fill=blue!20]
            (a) rectangle (c);           
    \end{pgfonlayer}
\end{tikzpicture}\vspace*{-0.5cm}
		\caption{Structure of the detector, which consists for each motion pattern of a recurrent neural network (LSTM) followed by a threshold to   obtain an output vector $\hbfz_n$ that consists of binary components. The hidden state $\bfh_n$ of the neural network  allows   the full structure to be recursive. \label{fig:detector}}\vspace*{-0.5cm}
	\end{figure}
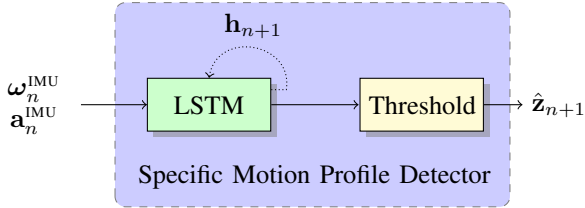

	The detector determines at each instant $n$ which ones of the specific motion profiles \eqref{eq:z} are valid, see Figure \ref{fig:detector}. The base core of the detector is a recurrent neural network, namely a Long-Short Term Memory (LSTM) \cite{goodfellowDeep2016}. The LSTMs take as input the IMU measurements and compute:
	\begin{align}
	\hbfu_{n+1},~ \bfh_{n+1} &= \mathrm{LSTM}\left(\left\lbrace\boomega^{\mathrm{\textsc{imu}}}_i,~ \bfa^{\mathrm{\textsc{imu}}}_i\right\rbrace_{i=0}^n \right) \\
	&= \mathrm{LSTM}\left(\boomega^{\mathrm{\textsc{imu}}}_n,~ \bfa^{\mathrm{\textsc{imu}}}_n,~ \bfh_n \right), \label{eq:lstm}
	\end{align}
	where $\hbfu_{n+1} \in \bbR^4$ contains probability scores for each motion profiles and $\bfh_{n}$ is the hidden state of the neural network. Probability scores are then converted to a binary vector $\hbfz_n = \mathrm{Threshold}\left(\hbfu_{n+1}\right)$ with a  threshold for each motion profile.

	The thresholds must be set with care, and the procedure will be described in  Section \ref{sec:implementation}.  Indeed, false alarms lead to degraded performance, since   a zero velocity assumption is incompatible with an actual motion. On the other hand, a missed profile is not harmful since it results in standard IMU based dead reckoning using \eqref{eq:prop1}-\eqref{eq:prop3}.

	\subsection{The Invariant Extended Kalman Filter}\label{sec:filter}
	
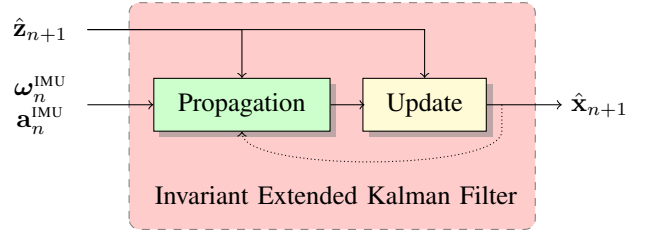
\begin{figure}
		\center
		\begin{tikzpicture}
    \node (wa) [draw, text width=6em,
    minimum height=2em, drop shadow, text centered,fill=green!20]  {Propagation};
    \path (wa.west)+(-1.5,0.0) node[text width=1cm,align=center] (asr2) {$\boomega_n^{\mathrm{\textsc{imu}}}$  $\bfa_n^{\mathrm{\textsc{imu}}}$};
    
    \path (wa.west)+(-1.5,1.0) node[text width=1cm,align=center] (z) {$\hbfz_{n+1}$};
    
    \path (wa.east)+(1.25,0.0) node (th) [draw, text width=4em,
    minimum height=2em, drop shadow, text centered,fill=yellow!20]  {Update};
    \path [draw, ->] (wa.east)+(0,0) -- ++(0.45,0);
    \path (th.east)+(1.5,0) node (vote)  {$\hbfx_{n+1}$};
	\path [draw,densely dotted, ->] (th.east)+(0.2,0) -- ++(0.2,-0.35) to[bend  right=-90,looseness=0.4, ->] (wa.south);   

    \path [draw, ->] (asr2.east) -- node [above] {} (wa.180);
    \path [draw, ->] (z) -| node [above] {} (wa.north);
    \path [draw, ->] (z) -| node [above] {} (th.north);
    \path [draw, ->] (th.east) -- node [above] {} (vote.west);
        
    \path (wa.south) +(1.25,-0.85) node (asrs) {Invariant Extended Kalman Filter};
    
    \begin{pgfonlayer}{background}
        \path (wa.south)+(-1.5,-1.3) node (a) {};
        \path (wa.north)+(3.9,1) node (c) {};
        \path[rounded corners, draw=black!50, dashed, fill=red!20]
            (a) rectangle (c);           
    \end{pgfonlayer}
\end{tikzpicture}\vspace*{-0.5cm}
		\caption{Structure of the IEKF. The filter leverages  motion profile information $\hbfz_n$ both for the propagation and the update  of the state $\hbfx_{n+1}$. \label{fig:filter}}\vspace*{-0.5cm}
	\end{figure}

	The extended Kalman filter is the most widespread technique in the  inertial navigation industry, since it was first successfully used in the Apollo program half a century ago. However, recent work advocates the use of a modified version, the  Invariant Extended Kalman Filter (IEKF) \cite{barrauInvariant2017} that has proved to bring drastic improvement over EKF, and has recently given raise to a commercial product, see \cite{barrauInvariant2018,barrauAligment2016}, and various successes in robotics \cite{wuinvariantEKF2017,brossardUnscented2018, heoConsistent2018, hartleyContactAided2018}. We thus opt for an IEKF to perform the fusion between the IMU measurements and the detected specific motion profiles. The IEKF outputs the state $\hbfx_{n}$ that consists of pose and velocity of the IMU, the IMU  biases,  along with their covariance.    We now describe the filter more in detail, whose architecture is displayed on Figure \ref{fig:filter}. 
	
	\subsubsection{IMU state}
	we define the IMU state  as 
	\begin{align}
	\bfx_n = \left(\bfR_n,~\bfv_n^{\mathrm{\textsc{w}}},~\bfp_n^{\mathrm{\textsc{w}}},~\bfb_{n}^{\boomega},~\bfb_{n}^{\bfa}\right), \label{eq:x}
	\end{align}
	which contains robot pose, velocity, and the IMU biases. The state evolution is given by 
	the dynamics \eqref{eq:prop1}-\eqref{eq:propend}, see Section \ref{sec:model}. As \eqref{eq:z_vel}-\eqref{eq:z_up} are all measurements expressed in the robot's frame, they lend themselves to the Right IEKF methodology, see \cite{barrauInvariant2017}. Applying it, we define the linearized state error as \begin{align}
	\bfe_n = \begin{bmatrix}\boxi_n \\ \bfe_n^{\bfb}\end{bmatrix} \sim \calN\left(\bfzero, \bfP_n\right),
	\end{align}
	where uncertainty is based on the use of the Lie exponential as advocated in  \cite{barfootAssociating2014} in a wheel odometry context, and mapped to the state as
	\begin{align}
	\bochi_n &= \exp_{SE_2(3)}\left(\boxi_n\right) \hbochi_n, \\
	\bfb_n &= \hbfb_n + \bfe_n^{\bfb},
	\end{align}
	where $\bochi_n \in SE_2(3)$ is a matrix that lives is the Lie group $SE_2(3)$ and represents the vehicle state $\bfR_n$, $\bfv_n^{\mathrm{\textsc{w}}}$, $\bfp_n^{\mathrm{\textsc{w}}}$ (see Appendix \ref{sec:se2_3} for the definition of $SE_2(3)$ and its exponential map), $\bfP_n \in \bbR^{15\times15}$ is the error state covariance matrix, $\bfb_n = \left(\bfb_{n}^{\boomega}, \bfb_{n}^{\bfa} \right) \in \bbR^6$ , $\hbfb_n = \left(\hbfb_{n}^{\boomega},\hbfb_{n}^{\bfa}\right) \in \bbR^6$, and $\hat{(\cdot)}$ denote estimated variables.
	
	\subsubsection{Propagation step}
	if no specific motion is detected, i.e. $\hz^{\mathrm{\textsc{vel}}}_{n+1}=0$, $\hz^{\mathrm{\textsc{ang}}}_{n+1}=0$, we apply \eqref{eq:prop1}-\eqref{eq:propend} to propagate the state and obtain $\hbfx_{n+1}$ and associated covariance through the Riccati equation\begin{align}
	\bfP_{n+1} = \bfF_n \bfP_n \bfF_n^T + \bfG_n \bfQ_n \bfG_n^T, \label{eq:covprop}
	\end{align}
	where the Jacobians $\bfF_n$, $\bfG_n$ are given in Appendix \ref{sec:jacobian},  and where $\bfQ_n$ denotes the covariance matrix of the noise $\bfw_n = \left(\bfw_{n}^{\boomega}, \bfw_{n}^{\bfa}, \bfw^{\bfb^{\boomega}}_{n}, \bfw^{\bfb^{\bfa}}_{n}\right) \sim \calN\left(\bfzero, \bfQ_n\right)$. By contrast, if a specific motion profile is detected, we modify model  \eqref{eq:prop1}-\eqref{eq:propend}  as follows:
	\begin{align}
	\hz_{n+1}^{\mathrm{\textsc{vel}}} = 1 \Rightarrow \left\lbrace\begin{matrix}
	\bfv_{n+1}^{\mathrm{\textsc{w}}} = \bfv_n^{\mathrm{\textsc{w}}} \\
	\bfp_{n+1}^{\mathrm{\textsc{w}}} = \bfp_n^{\mathrm{\textsc{w}}}
	\end{matrix}\right., \label{eq:prop4} \\
	\hz_{n+1}^{\mathrm{\textsc{ang}}} = 1 \Rightarrow \bfR_{n+1} = \bfR_n, \label{eq:prop5}
	\end{align}and the estimated state $\hbfx_{n+1}$ and covariance $\bfP_{n+1}$ are modified accordingly.

	\subsubsection{Update} 
	each motion profile yields one of the following pseudo-measurements:
	\begin{align}
	\bfy_{n+1}^{\mathrm{\textsc{vel}}} &= \begin{bmatrix}
	\bfR_{n+1}^T\bfv_{n+1}^{\mathrm{\textsc{w}}} \\
	\bfb_{n+1}^{\bfa} -\bfR_{n+1}^T\bfg
	\end{bmatrix} = \begin{bmatrix}\bfzero \label{eq:y_vel}\\ \bfa_{n}^{\mathrm{\textsc{imu}}}\end{bmatrix}, \\
	\bfy_{n+1}^{\mathrm{\textsc{ang}}} &= \bfb_{n+1}^{\boomega} = \boomega_{n}^{\mathrm{\textsc{imu}}}, \label{eq:y_ang}\\
	\bfy_{n+1}^{\mathrm{\textsc{lat}}} &= v_{n+1}^{\mathrm{\textsc{lat}}} = 0, \label{eq:y_lat}\\
	\bfy_{n+1}^{\mathrm{\textsc{up}}} &= v_{n+1}^{\mathrm{\textsc{up}}} = 0  \label{eq:y_up}.
	\end{align}
	A vector $\bfy_{n+1}$ is computed by stacking the pseudo-measurements of the detected motion profiles. Note that, if $\hz_{n+1}^{\mathrm{\textsc{vel}}} = 1$ we do not consider \eqref{eq:y_lat}-\eqref{eq:y_up} since \eqref{eq:y_vel} implies \eqref{eq:y_lat}-\eqref{eq:y_up}. If no specific motion is detected, the update step is skipped, otherwise we follow the IEKF methodology \cite{barrauInvariant2017} and compute
	\begin{align}
	\bfK &= \bfP_{n+1} \bfH_{n+1}^T/\left(\bfH_{n+1} \bfP_{n+1} \bfH_{n+1}^T + \bfN_{n+1}\right), \label{eq:gain}\\
	\bfe^+ &= \bfK\left(\bfy_{n+1}-\hbfy_{n+1}\right) = \begin{bmatrix}
	\boxi^+ \\ \bfe^{\bfb+}
	\end{bmatrix}, \label{eq:innovation} \\
	\hbochi_{n+1}^+ &= \exp\left(\boxi^+\right) \hbochi_{n+1},~ \bfb_{n+1}^+ = \bfb_{n+1} + \bfe^{\bfb+}, \label{eq:upstate}\\
	\bfP_{n+1}^+ &= \left(\bfI - \bfK \bfH_{n+1}\right) \bfP_{n+1}, \label{eq:upcov}
	\end{align}
	summarized as Kalman gain \eqref{eq:gain}, state innovation \eqref{eq:innovation}, state update \eqref{eq:upstate} and covariance update \eqref{eq:upcov}, where $\bfH_{n+1}$ is the measurement Jacobian matrix given in Appendix \ref{sec:jacobian} and $\bfN_{n+1}$ the noise measurement covariance.
	
	\subsubsection{Initialization} launching the system when the platform is moving without estimating biases and orientation can induce drift at the beginning which then is impossible to compensate.  As the filter is able to correctly self-initialize the biases, pitch, roll and its  covariance matrix  when the trajectory is first stationary, we enforce each sequence in  Section \ref{sec:results} to start by a minimum of \SI{1}{s} stop. Admittedly restrictive, the required stop is of extremely short duration, especially as compared to standard calibration techniques \cite{safranInertial2018}.

	\section{Results on Car Dataset}\label{sec:results}
\begin{figure}
		\centering
		\includegraphics[height=3.0cm]{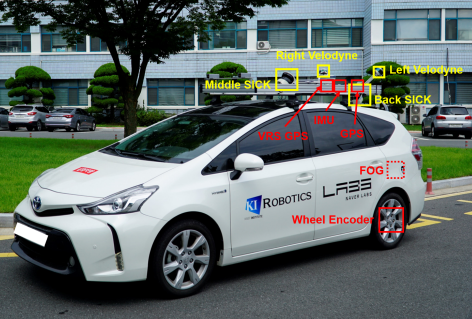}
		\includegraphics[height=3.0cm]{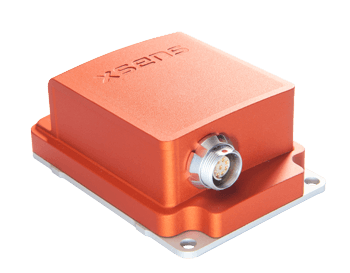}

		\caption{\label{fig:dataset} The considered dataset \cite{jeongComplex2018} contains data logs of a Xsens MTi-300\protect\footnotemark ~(right) recorded at \SI{100}{Hz} along with the ground truth pose.\vspace*{-0.3cm}}
	\end{figure}

	The following results are obtained on the \emph{complex urban LiDAR dataset} \cite{jeongComplex2018}, that consists of data recorded on a consumer car moving in complex urban environments, e.g. metropolitan  areas,  large  building  complexes  and  underground parking lots, see Figure \ref{fig:dataset}. Our goal is to show that using an IMU of moderate cost, one manages to obtain surprisingly accurate dead reckoning by using state-of-the-art machine learning techniques to detect relevant assumptions that can be fed into a state-of-the-art Kalman filter. The detector is trained on a series of sequences and tested on another sequences, but all sequences involve the same car and inertial sensors. Generalization to different robot and IMU is not considered herein and is left for future work.
	
	\setcounter{footnote}{0}
	\footnotetext{\url{https://www.xsens.com/}}
	
	\subsection{Implementation Details}\label{sec:implementation}
	We provide in this section the detector and filter setting of the RINS-W system. The detector disposes of four LSTMs, one for each motion profile.  Each LSTM consists of 2 layers of 250 hidden units and its hidden state  is mapped to a score probability by a 2 layers multi-perceptron network with a ReLU activation function and is followed by a sigmoid function \cite{goodfellowDeep2016} that outputs a scalar value in the range $\left[0,~1\right]$. We implement the detector on PyTorch\footnote{\url{https://pytorch.org/}} and set the threshold values to 0.95 for ($z_n^{\mathrm{\textsc{vel}}}$,  $z_n^{\mathrm{\textsc{ang}}}$), and 0.5 for ($z_n^{\mathrm{\textsc{lat}}}$, $z_n^{\mathrm{\textsc{up}}}$). The filter operates at the \SI{100}{Hz} IMU rate ($dt=\SI{e-2}{s}$) and its noise covariance matrices are parameterized as
	\begin{align}
	\bfQ_n &= \diag\left(\sigma_{\boomega}^2\bfI,~\sigma_{\bfa}^2\bfI,~ \sigma_{\bfb_{\boomega}}^2\bfI,~ \sigma_{\bfb_{\bfa}}^2\bfI\right), \\
	\bfN_n &= \diag\left(\sigma_{\mathrm{\textsc{vel}, \bfv}}^2\bfI,~\sigma_{\mathrm{\textsc{vel}, \bfa}}^2\bfI,~ \sigma_{\mathrm{\textsc{ang}}}^2\bfI,~ \sigma_{\mathrm{\textsc{lat}}}^2,~ \sigma_{\mathrm{\textsc{up}}}^2 \right),
	\end{align}
	where we set $\sigma_{\boomega} = \SI{0.01}{rad/s}$,  $\sigma_{\bfa} = \SI{0.2}{m/s^2}$, $\sigma_{\bfb_{\boomega}} = \SI{0.001}{rad/s}$, $\sigma_{\bfb_{\bfa}} = \SI{0.02}{m/s^2}$ for the noise propagation covariance matrix $\bfQ_n$, and $\sigma_{\mathrm{\textsc{vel}, \bfv}} = \SI{1}{m/s}$, $\sigma_{\mathrm{\textsc{vel}, \bfa}} = \SI{0.4}{m/s^2}$,
	$\sigma_{\mathrm{\textsc{ang}}} = \SI{0.04}{rad/s}$, $\sigma_{\mathrm{\textsc{lat}}} = \SI{3}{m/s}$, and $\sigma_{\mathrm{\textsc{up}}} = \SI{3}{m/s}$ for the noise measurement covariance matrix $\bfN_n$.

	\subsection{Detector Training}\label{sec:train}
	The detector is trained with the sequences \texttt{urban06} to \texttt{urban14}, that represents  \SI{100}{km} of training data (sequences \texttt{urban00} to \texttt{urban05} does not have acceleration data). For each sequence, we compute ground truth position velocity $\bfv_n^{\mathrm{\textsc{w}}}$ and angular  velocity $\boomega_n$ after differentiating the ground pose and applying smoothing. We then compute the ground-truth $\bfz_n$ by applying a small threshold on the ground truth velocities, e.g. we consider $z_n^{\mathrm{\textsc{vel}}}=1$ if $\|\bfv_n^{\mathrm{\textsc{w}}} \| < \SI{0.01}{m/s}$. We set similarly the other motion profiles and use a threshold of \SI{0.005}{rad/s} for the angular velocity, and a threshold of \SI{0.1}{m/s} for the lateral and upward velocities.
	
	The detector is trained during 500 epochs with the ADAM optimizer \cite{kingmaAdam2014}, whose learning rate is initializing at $10^{-3}$ and managed by a learning rate scheduler. Regularization is enforced with dropout layer, where $p=0.4$  is the probability of any element to be zero. We use the binary cross entropy loss since we have four binary classification problems. For each epoch, we organize data as a batch of \SI{2}{min} sequences, where we randomly set the start instant of each sequence, and constraints each starting sequence to be a stop of at minimum  \SI{1}{s}. Training the full detector takes less than one day with a GTX 1080 GPU.

	\renewcommand{\figurename}{Table}
	\setcounter{figure}{0} 
	\begin{figure}
		\centering
		\begin{tabular}{c||c|c|c}
			\toprule
			test seq. & wheels odo. & odo. + FoG & \textbf{RINS-W}   \\
			\midrule
			\small 15: \SI{16}{min} & \small $19$ / $5$ / $36$ & \small $\textbf{7}$ / $\textbf{2}$ /  $\textbf{7}$ & \small $\mathbf{7}$ / $5$ / $12$ \\
			\small 16: \SI{73}{min} & \scriptsize $140$ / $127$ / $1166$ & \small $34$ / $20$ / 164 & \small $\mathbf{27}$ / $\mathbf{11}$ / $\mathbf{20}$  \\
			\small 17: \SI{19}{min} & \small $96$ / $64$ / $427$ & \small $58$ / $51$ / $166$ & \small $\mathbf{13}$ / $\mathbf{11}$ / $\mathbf{13}$  \\
			\midrule
			\small 15-17: & \multirow{2}{*}{\small $114$ / $98$ / $677$} & \multirow{2}{*}{\small $34$ / $30$ / $152$} & \multirow{2}{*}{\small $\mathbf{22}$ / $\mathbf{10}$ / $\mathbf{18}$} \\
			\small \SI{108}{min} & & & \\
			\bottomrule
		\end{tabular}
		\caption{Results obtained by the 3 methods on \texttt{urban} test sequences  15, 16, 17 in terms of: $m$-ATE /aligned $m$-ATE / final distance error to ground truth, in \SI{}{m}.  Last line is the concatenation of the three sequences. Direct IMU integration   always diverges. The proposed RINS-W outperforms differential wheel speeds based odometry and  outperforms on average (see last line) the expensive combination of odometry +   FoG. Indeed, RINS-W uses an IMU with gyro stability of \SI{10}{\deg/h}, whereas FoG stability is \SI{0.05}{\deg/h}. \label{fig:final_error}} 
	\end{figure}
	\renewcommand{\figurename}{Fig.}
	\setcounter{figure}{6} 

	\subsection{Evaluation Metrics}
	To assess performances   we consider three error metrics: 
	\subsubsection{Mean Absolute Trajectory Error ($m$-ATE)} which averages the planar translation error of estimated poses with respect to a ground truth trajectory and is less sensitive to single poor estimates than root mean square error;
	\subsubsection{Mean Absolute Aligned Trajectory Error (aligned $m$-ATE)} that first aligns the estimated trajectory with the ground truth and then computes the mean absolute trajectory error. This metric evaluates the consistency of the trajectory estimates;
	\subsubsection{Final distance error} which is the final distance between the un-aligned estimates and the ground truth.
	
	\subsection{Trajectory Results}
	
	After training the detector on  sequences \texttt{urban06} to \texttt{urban14}, we evaluate the   approach on test sequences \texttt{urban15} to \texttt{urban17}, that represent \SI{40}{km} of evaluation data. We compare 4 methods:
	\begin{itemize}
		\item \textbf{IMU}: the direct integration of the IMU measurements based on \eqref{eq:prop1}-\eqref{eq:propend}, that is, pure inertial navigation;
		\item \textbf{Odometry}: the integration of a differential wheel encoder which computes linear and angular velocities;
		\item \textbf{RINS-W (ours)}: the proposed approach, that uses only the IMU signals and involves no other sensor. 
		\item \textbf{Odometry + FoG}: the integration of a differential wheel encoder which computes only linear velocity. The angular velocity is obtained after integrating the increments of an highly accurate and costly KVH DSP-1760\footnote{\url{https://www.kvh.com}} Fiber optics Gyro (FoG). The FoG gyro bias stability  (\SI{0.05}{\deg/h}) is 200 times smaller than the gyro stability of the IMU used by RINS-W;
	\end{itemize}

	\begin{figure}
		
		\begin{tikzpicture}[]
		\begin{axis}[height=9.5cm,
		width=8.5cm,
		ylabel=$y$ (\SI{}{km}),
		xlabel=$x$ (\SI{}{km}),
		xmax=0.08,
		xmin=-0.78,
		ymax=0.75,
		ymin=-0.3,
		ylabel style={xshift=0.0cm,yshift=-0.0cm},
		xlabel style={xshift=0cm,yshift=0cm},
		legend columns=2, legend pos= north east, 
		label style={font=\small}, 
		ticks=both,
		legend entries={ground truth, IMU, odometry, odometry+FoG, \textbf{proposed RINS-W}},
		legend style={font=\scriptsize}, 
		yticklabel style={
			/pgf/number format/fixed,
			/pgf/number format/precision=2,
			font=\small
		},
		xticklabel style={font=\small,
			/pgf/number format/fixed,
			/pgf/number format/precision=2,}
		]
		
		\addplot[draw = black, 
		very thick] table[x=x,y=y] {figures/urban15gt.txt};
		\addplot[draw = cyan, dashed, thick] table[x=x,y=y] {figures/urban15imu.txt};
		\addplot[draw = red, dashed, thick] table[x=x,y=y] {figures/urban15odo.txt};
		\addplot[draw = blue, dashed, thick] table[x=x,y=y] {figures/urban15odofog.txt};
		\addplot[draw = green, dashed, thick] table[x=x,y=y] {figures/urban15hat.txt};
		\node at (-0.01,0.15) {\Large \color{red}\textbf{*}};
		\node at (-0.04,0.17) {\Large \color{blue}\textbf{*}};
		\node at (-0.016,0.142) {\Large \color{green}\textbf{*}};
		\path [draw, ->] (-0.15, 0.08)  node [left, text width=1.2cm, text centered] {\small ground truth end} to[bend  right=-40,looseness=1.1, ->] (-0.04, 0.15);
		\path [draw, ->] (-0.13, -0.03)  node [below] {start} to[bend  right=-40,looseness=1.3, ->] (-0.01, 0.01);
		\end{axis}	
		\end{tikzpicture}
		\vspace*{-0.3cm}
		\caption{Ground truth and trajectory estimates for the test sequence \texttt{urban15} of the car dataset \cite{jeongComplex2018}. RINS-W obtains  results comparable with FoG-based odometry. \label{fig:urban15}\vspace*{-0.3cm}}
	\end{figure}
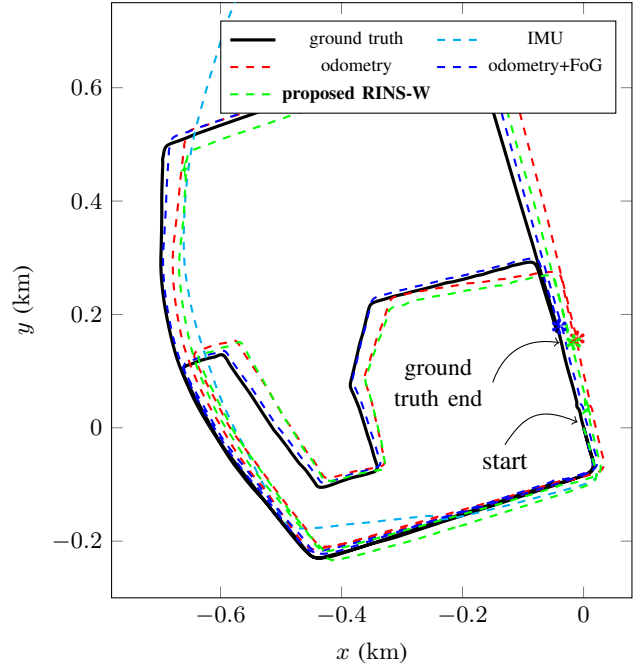
	
	We delay each sequence such that the trajectory starts with   a \SI{1}{s}   stop to initialize the orientation and the biases, see Section \ref{sec:filter}. Bias initialization is \emph{also} performed for  IMU pure integration and the FoG. Optimized parameters for wheel speeds sensors  calibration are provided  by the dataset.
	
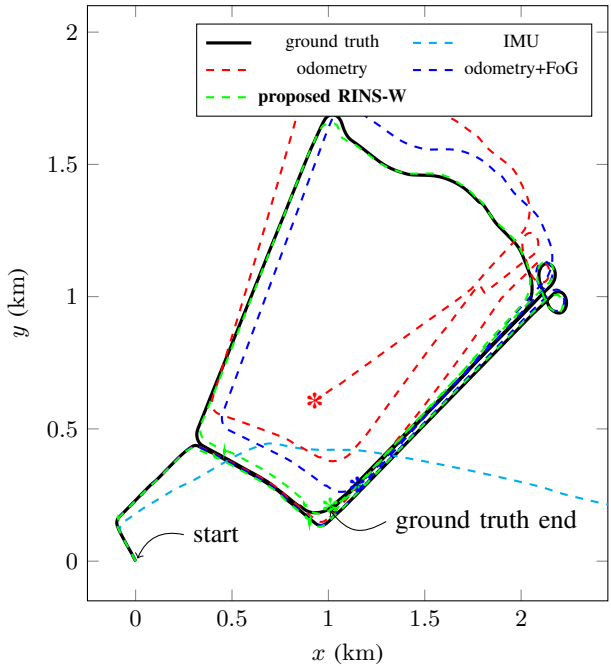
\begin{figure}
		\begin{tikzpicture}[]
		\begin{axis}[height=9.5cm,
		width=8.5cm,
		ylabel=$y$ (\SI{}{km}),
		xlabel=$x$ (\SI{}{km}),
		xmax=2.45,
		xmin=-0.25,
		ymax=2.1,
		ymin=-0.15,
		ylabel style={xshift=0.0cm,yshift=-0.0cm},
		xlabel style={xshift=0cm,yshift=0cm},
		legend columns=2, legend pos= north east, 
		label style={font=\small}, 
		ticks=both,
		legend entries={ground truth, IMU, odometry, odometry+FoG, \textbf{proposed RINS-W}},
		legend style={font=\scriptsize}, 
		yticklabel style={
			/pgf/number format/fixed,
			/pgf/number format/precision=2,
			font=\small
		},
		xticklabel style={font=\small,
			/pgf/number format/fixed,
			/pgf/number format/precision=2,}
		]
		
		\addplot[draw = black, 
		very thick] table[x=x,y=y] {figures/urban17gt.txt};
		\addplot[draw = cyan, dashed, thick] table[x=x,y=y] {figures/urban17imu.txt};
		\addplot[draw = red, dashed, thick] table[x=x,y=y] {figures/urban17odo.txt};
		\addplot[draw = blue, dashed, thick] table[x=x,y=y] {figures/urban17odofog.txt};
		\addplot[draw = green, dashed, thick] table[x=x,y=y] {figures/urban17hat.txt};
		\node at (0.93,0.59) {\Large \color{red}\textbf{*}};
		\node at (1.15,0.27) {\Large \color{blue}\textbf{*}};
		\node at (1.01,0.194) {\Large \color{green}\textbf{*}};
		\path [draw, ->] (1.3, 0.15)  node [right] {ground truth end} to[bend  right=-40,looseness=1.3, ->] (1.01, 0.19);
		\path [draw, ->] (0.25, 0.1)  node [right] {start} to[bend  left=-40,looseness=1.1, ->] (0.01, 0.01);
		\end{axis}	
		\end{tikzpicture}\vspace*{-0.2cm}
		\caption{Ground truth and trajectory estimates for the test sequence \texttt{urban17} of the car dataset \cite{jeongComplex2018}. RINS-W clearly outperforms the odometry and even the odometry + FoG solution. We note that RINS-W accurately follows the interchange road located at $(x=2, y=1)$. \label{fig:urban17}\vspace*{-0.2cm}}
	\end{figure}

	Experimental results in terms of error with respect to ground truth are displayed in Table \ref{fig:final_error}, and illustrated on  Figures \ref{fig:results},  \ref{fig:urban15}, and   \ref{fig:urban17}. Results demonstrate that:
	\begin{itemize}
		\item directly integrating IMU leads to divergence at the first turn, even after a careful calibration procedure;
		\item wheel-based differential odometry accurately estimates the linear velocity but has troubles estimating the   yaw   during sharp bends, even if the dataset has been obtained in an urban environment and the odometry parameters are calibrated. This  drawback  may  be remedied at the expense of using an additional high-cost gyroscope;
		\item the proposed scheme completely outperforms wheel encoders, albeit in urban environment. More surprisingly, our approach competes with the combination of  wheel speed sensors + (200 hundred times more accurate) Fyber optics Gyro, and even outperforms it on average.
	\end{itemize}
	Furthermore, although comparisons were performed in 2D environments  our method yields the full 3D pose of the robot, and as such is compatible with non planar environments.

	\subsection{Discussion}\label{sec:resbonus}
	The performances of RINS-W can be explained by: $i$) a   false-alarm free detector; $ii$) the fact incorporating side information into IMU integration obviously yields better results; and $iii$)  the use of a recent IEKF that has been proved to be well suited for localization tasks. 
	
\begin{figure}
		\begin{tikzpicture}
		\begin{axis}[height=9.5cm,
		width=8.5cm,
		ylabel=$y$ (\SI{}{km}),
		xlabel=$x$ (\SI{}{km}),
		xmax=0.27,
		xmin=-0.06,
		ymax=0.39,
		ymin=-0.02,
		ylabel style={xshift=0.0cm,yshift=-0.0cm},
		xlabel style={xshift=0cm,yshift=0cm},
		legend columns=1, legend pos= north east, 
		legend entries={true trajectory, usual estimation, \textbf{proposed estimation}},
		yticklabel style={
			/pgf/number format/fixed,
			/pgf/number format/precision=2,
		},
		xticklabel style={
			/pgf/number format/fixed,
			/pgf/number format/precision=2,}
		]
		
		\addplot[draw = black, 
		very thick] table[x=x,y=y] {figures/07gt.txt};
		\addplot[draw = orange, dotted, thick] table[x=x,y=y] {figures/7imu.txt};
		\addplot[draw = green, dashed, thick] table[x=x,y=y] {figures/07hat.txt};
		\path [draw, ->] (-0.028, 0.023)  node [above] {start} to[bend  left=-40,looseness=1., ->] (-0.005, 0);
		\end{axis}	
		\end{tikzpicture}\vspace*{-0.2cm}
		\caption{Comparison on the sequence \texttt{urban07}  between proposed RINS-W, and a similar algorithm that does not   use   zero lateral and vertical velocities   assumptions brought by the detector, see  \eqref{eq:y_lat}-\eqref{eq:y_up}. The final distance between ground truth and RINS-W estimates is as small as \SI{5}{m}, whereas ignoring \eqref{eq:y_lat}-\eqref{eq:y_up} yields divergence. \label{fig:seq77}\vspace*{-0.2cm}}
	\end{figure}
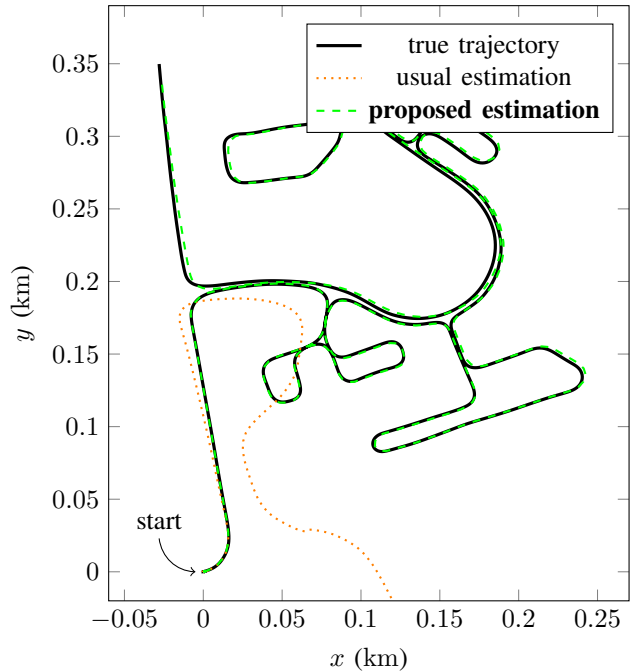

	We also emphasize the importance of  \eqref{eq:z_lat} and \eqref{eq:z_up} in the  procedure, i.e. applying \eqref{eq:y_lat}-\eqref{eq:y_up}. For illustration, we consider sequence  \texttt{urban07} of   \cite{jeongComplex2018}, where the vehicle moves during 7 minutes without stop so that ZUPT may not be used. We implement  the detector  trained on the first 6 sequences, and   compare   the proposed RINS-W to a RINS-W which does \emph{not}  use updates \eqref{eq:y_lat}-\eqref{eq:y_up} when detected, see  Figure \ref{fig:seq77}.  Clearly, the reduced RINS-W diverges at the first turn whereas the full RINS-W is accurate along all the trajectory and obtains a final distance w.r.t. ground truth of \SI{5}{m}. In contrast, odometry + FoG achieves  \SI{16}{m}.

	\subsection{Detector Evaluation}\label{sec:detector_results}
	In Section \ref{sec:resbonus} we mentioned  three  possible reasons explaining the performances of RINS-W, but could not assess what is owed to each. To assess the detector's performance, and to demonstrate the interest of  our powerful deep neural network based approach (see Section \ref{sec:detector}) we   focus on the zero velocity detection \eqref{eq:z_vel}, and compare the detector with the Acceleration-Moving Variance Detector (AMVD) \cite{skogZeroVelocity2010} on the test sequences \texttt{urban15-17}, which represent $64.10^4$ measurements. The AMVD computes the accelerometer variance over a sample window $W=10^2$ and assumes the vehicle is stationary if the variance  falls below a threshold $\gamma=10^{-3}$. To make sure AMVD performs at its best, the parameters $W$ and $\gamma$ are optimized by grid search \emph{on the test sequences}. Results are shown in Table \ref{fig:detector_result} and demonstrate  the detector is more accurate than this ``ideal''  AMVD.

    \renewcommand{\figurename}{Table}
	\setcounter{figure}{1} 
	\begin{figure}
		\centering
		\begin{tabular}{c||c|c}
			\toprule
			$z_n^{\mathrm{vel}}$ detection & ideal AMVD &  {our detector} \\
			\midrule
			\small true positive / false pos. & \small $47.10^4$ / $4.10^3$ & \small $\mathbf{48.10^4}$ / $\mathbf{7.10^2}$  \\
			\small true negative / false neg. & \small $\mathbf{16.10^4}$ / $1.10^4$ & \small $\mathbf{16.10^4}$ / $\mathbf{9.10^3}$  \\  
			\small precision / recall & \small $0.974$ / $\mathbf{0.940}$ & \small $\mathbf{0.996}$ / $\mathbf{0.940}$ \\
			\bottomrule
		\end{tabular}
		\caption{Results on zero velocity \eqref{eq:z_vel} detection obtained by an  \emph{ideal} AMVD   \cite{skogZeroVelocity2010} and the proposed detector on test sequences \texttt{urban15-17}. The proposed detector systematically obtains better results and precision. This is remarkable because  the detector is {not trained} on those sequences, whereas   AMVD parameters were  {optimized  on the considered test sequences}.\label{fig:detector_result}\vspace*{-0.1cm}} 
	\end{figure}
	\renewcommand{\figurename}{Fig.}
	\setcounter{figure}{9} 
	

	\section{Conclusion}\label{sec:con}
	This paper proposes a novel approach for robust inertial navigation for wheeled robots (RINS-W). Even if an autonomous vehicle is equipped with wheel encoders, LiDAR, and vision besides the IMU, the algorithm may be run in parallel for safety, in case of sensor failure, or more simply for validation such as slip angle monitoring \cite{oxts}. Our approach exploits deep neural networks to identify specific patterns in wheeled vehicle motions and  incorporates this knowledge in IMU integration for localization. The entire algorithm is fed with IMU signals only, and requires no other sensor. The method leads to surprisingly accurate results, and opens  new perspectives for combination of data-driven methods with well established methods for autonomous systems.  Moreover, the pseudo measurements output by the detector are not reserved for dead reckoning and may prove useful in any fusion scheme. In future work, we would like to address the   learning of the Kalman covariance matrices, and also the issue of generalization from one vehicle to another. 
	
	\appendices
	\section{}
	\label{sec:se2_3}
	The Lie group $SE_2(3)$ is an extension of the Lie group $SE(3) $ and is described as follow, see  \cite{barrauInvariant2017} for more details. A $5\times5$ matrix $\bochi_n \in SE_{2}(3)$ is defined as
	\begin{align}
	\bochi_n =  \left[\begin{array}{ccc}
	\bfR_n & \bfv_n^{\mathrm{\textsc{w}}} & \bfp_n^{\mathrm{\textsc{w}}} \\
	\bfzero & \multicolumn{2}{c}{\bfI}
	\end{array} \right] \in SE_2(3).\label{bochi:eq}
	\end{align}
	The uncertainties $\boxi_n \in \bbR^{9}$ are mapped to the Lie algebra $\mathfrak{se}_2(3)$ through the transformation $\boxi_n\mapsto\boxi_n^{\wedge} $ defined as
	\begin{align}
	\boxi_n &= \left(\boxi^\bfR_n,~ \boxi^\bfv_n,~ \boxi^\bfp_n \right), \\
	\boxi^{\wedge}_n &= \left[\begin{array}{ccc} \left(\boxi^\bfR_n\right)_\times & \boxi^\bfv_n & \boxi^\bfp_n  \\
	\multicolumn{3}{c}{\bfzero}
	\end{array} \right] \in \mathfrak{se}_2(3),
	\end{align}
	where $(\cdot)_\times$ transforms a vector to a skew-symmetric matrix, $\boxi^\bfR_n \in \bbR^3$, $\boxi^\bfv_n \in \bbR^3$ and $\boxi^\bfp_n \in \bbR^3$. The closed-form expression for the exponential map is given as
	\begin{align}
	\exp_{SE_2(3)}\left(\boxi_n\right) = \bfI + \boxi_n^\wedge + a (\boxi_n^{\wedge})^{2} + b (\boxi_n^{\wedge})^{3},
	\end{align}
	where $a=\frac{1-\cos\left(\|\boxi^\bfR_n\|\right)}{\|\boxi^\bfR_n\|}$ and $b=\frac{ \|\boxi^\bfR_n\|-\sin\left(\|\boxi^\bfR_n\|\right)}{\|\boxi^\bfR_n\|^3}$.
	
	\section{}
	\label{sec:jacobian}
	Following the Right IEKF of \cite{barrauInvariant2017}, the Jacobians required for the computation of the filter propagation \eqref{eq:covprop} are given as
	\begin{align}
	\bfF_n = \bfI + \begin{bmatrix}
	\bfzero & \bfzero & \bfzero & -\bfR_n & \bfzero \\
	(\bfg)_\times & \bfzero & \bfzero & -(\bfv_n^{\mathrm{\textsc{w}}})_\times\bfR_n & -\bfR_n  \\
	\bfzero & \bfI & \bfzero & -(\bfp_n^{\mathrm{\textsc{w}}})_\times\bfR_n & \bfzero \\
	\bfzero & \bfzero & \bfzero & \bfzero & \bfzero \\
	\bfzero & \bfzero & \bfzero & \bfzero & \bfzero
	\end{bmatrix} dt, \label{eq:F}
	\end{align}
	\begin{align}
	\bfG_n = \begin{bmatrix}
	\bfR_n & \bfzero & \bfzero &\bfzero \\
	(\bfv_n^{\mathrm{\textsc{w}}})_\times \bfR_n & \bfR_n & \bfzero &\bfzero \\
	(\bfp_n^{\mathrm{\textsc{w}}})_\times \bfR_n & \bfzero & \bfzero &\bfzero \\
	\bfzero & \bfzero & \bfI &\bfzero \\
	\bfzero & \bfzero & \bfzero &\bfI 
	\end{bmatrix} dt,
	\end{align}
	when $\hz_n^{\mathrm{\textsc{vel}}} =0$ and $\hz_n^{\mathrm{\textsc{ang}}} =0$. Otherwise, we set the appropriate rows to  zero in $\bfF_n$ and $\bfG_n$, i.e.:
	\begin{itemize}
		\item if $\hz_n^{\mathrm{\textsc{vel}}} =1$ we set the 4 to 9 rows of the right part of $\bfF_n$ in \eqref{eq:F} and of $\bfG_n$ to zero.
		\item if $\hz_n^{\mathrm{\textsc{ang}}} =1$ we set the 3 first rows of the right part of $\bfF_n$ in \eqref{eq:F} and of $\bfG_n$ to zero.
	\end{itemize}
	
	Once again following \cite{barrauInvariant2017}, the measurement Jacobians used in  the filter update \eqref{eq:gain}-\eqref{eq:upcov} are given as
	\begin{align}
	\bfH_n^{\mathrm{\textsc{vel}}} &= \begin{bmatrix}
	\bfzero & \bfR_n^T & \bfzero & \bfzero & \bfzero \\
	\bfR_n^T (\bfg)_\times & \bfzero & \bfzero & \bfzero & -\bfI  
	\end{bmatrix},  \\
	\bfH_n^{\mathrm{\textsc{ang}}} &= \begin{bmatrix}
	\bfzero & \bfzero & \bfzero & -\bfI & \bfzero 
	\end{bmatrix},
	\end{align}
	and we obtains $\bfH_n^{\mathrm{\textsc{lat}}}$ and $\bfH_n^{\mathrm{\textsc{up}}}$ as respectively the second and third row of $\bfH_n^{\mathrm{\textsc{vel}}}$.
	\bibliographystyle{IEEEtran}
	\bibliography{biblio}

\begin{thebibliography}{10}
\providecommand{\url}[1]{#1}
\csname url@samestyle\endcsname
\providecommand{\newblock}{\relax}
\providecommand{\bibinfo}[2]{#2}
\providecommand{\BIBentrySTDinterwordspacing}{\spaceskip=0pt\relax}
\providecommand{\BIBentryALTinterwordstretchfactor}{4}
\providecommand{\BIBentryALTinterwordspacing}{\spaceskip=\fontdimen2\font plus
\BIBentryALTinterwordstretchfactor\fontdimen3\font minus
  \fontdimen4\font\relax}
\providecommand{\BIBforeignlanguage}[2]{{%
\expandafter\ifx\csname l@#1\endcsname\relax
\typeout{** WARNING: IEEEtran.bst: No hyphenation pattern has been}%
\typeout{** loaded for the language `#1'. Using the pattern for}%
\typeout{** the default language instead.}%
\else
\language=\csname l@#1\endcsname
\fi
#2}}
\providecommand{\BIBdecl}{\relax}
\BIBdecl

\bibitem{oxts}
\BIBentryALTinterwordspacing
OxTS. (2018) {W}hy it is {N}ecessary to {I}ntegrate an {I}nertial {M}easurement
  {U}nit with {I}maging {S}ystems on an {A}utonomous {V}ehicle. [Online].
  Available:
  \url{https://www.oxts.com/technical-notes/why-use-ins-with-autonomous-vehicle/}
\BIBentrySTDinterwordspacing

\bibitem{collinInertial2018}
J.~Collin, P.~Davidson, M.~Kirkko-Jaakkola, and H.~Leppäkoski, ``Inertial
  {{Sensors}} and {{Their Applications}},'' in \emph{Handbook of {{Signal
  Processing Systems}}}.\hskip 1em plus 0.5em minus 0.4em\relax {Springer},
  2018, pp. 69--96.

\bibitem{kokUsing2017}
M.~Kok, J.~Hol, and T.~Schön, ``Using {{Inertial Sensors}} for {{Position}}
  and {{Orientation Estimation}},'' \emph{Foundations and Trends® in Signal
  Processing}, vol.~11, no. 1-2, pp. 1--153, 2017.

\bibitem{safranInertial2018}
\BIBentryALTinterwordspacing
Safran. (2018) Inertial navigation systems. [Online]. Available:
  \url{https://www.safran-group.com/fr/video/13612}
\BIBentrySTDinterwordspacing

\bibitem{cadenaPresent2016}
C.~Cadena, L.~Carlone, H.~Carrillo, Y.~Latif, D.~Scaramuzza, J.~Neira, I.~Reid,
  and J.~Leonard, ``Past, {{Present}}, and {{Future}} of {{Simultaneous
  Localization}} and {{Mapping}}: {{Toward}} the {{Robust}}-{{Perception
  Age}},'' \emph{IEEE T-RO}, pp. 1309--1332, 2016.

\bibitem{sunRobust2018}
K.~Sun, K.~Mohta, B.~Pfrommer, M.~Watterson, S.~Liu, Y.~Mulgaonkar, C.~J.
  Taylor, and V.~Kumar, ``Robust {{Stereo Visual Inertial Odometry}} for {{Fast
  Autonomous Flight}},'' \emph{IEEE RA-L}, vol.~3, no.~2, pp. 965--972, 2018.

\bibitem{deschaudIMLSSLAM2018}
J.-E. Deschaud, ``{{IMLS}}-{{SLAM}}: {{Scan}}-to-{{Model Matching Based}} on
  {{3D Data}},'' in \emph{IEEE ICRA}, 2018.

\bibitem{goodfellowDeep2016}
I.~Goodfellow, Y.~Bengio, and A.~Courville, \emph{Deep {{Learning}}}.\hskip 1em
  plus 0.5em minus 0.4em\relax {The MIT press}, 2016.

\bibitem{barrauInvariant2017}
A.~Barrau and S.~Bonnabel, ``The {{Invariant Extended Kalman Filter}} as a
  {{Stable Observer}},'' \emph{IEEE Trans. on Automatic Control}, vol.~62,
  no.~4, pp. 1797--1812, 2017.

\bibitem{barrauInvariant2018}
\BIBentryALTinterwordspacing
------, ``Invariant {{Kalman Filtering}},'' \emph{Annual Review of Control,
  Robotics, and Autonomous Systems}, vol.~1, no.~1, pp. 237--257, 2018.
  [Online]. Available:
  \url{https://doi.org/10.1146/annurev-control-060117-105010}
\BIBentrySTDinterwordspacing

\bibitem{jeongComplex2018}
J.~Jeong, Y.~Cho, Y.-S. Shin, H.~Roh, and A.~Kim, ``Complex {{Urban LiDAR Data
  Set}},'' in \emph{IEEE ICRA}, 2018.

\bibitem{barrauAligment2016}
A.~Barrau and S.~Bonnabel, ``Aligment {{Method}} for an {{Inertial Unit}},''
  patent 15/037,653, 2016.

\bibitem{wuinvariantEKF2017}
K.~Wu, T.~Zhang, D.~Su, S.~Huang, and G.~Dissanayake, ``An invariant-{{EKF
  VINS}} algorithm for improving consistency,'' in \emph{{IEEE IROS}}, 2017,
  pp. 1578--1585.

\bibitem{heoConsistent2018}
S.~Heo and C.~G. Park, ``Consistent {{EKF}}-{{Based Visual}}-{{Inertial
  Odometry}} on {{Matrix Lie Group}},'' \emph{IEEE Sensors Journal}, vol.~18,
  no.~9, pp. 3780--3788, 2018.

\bibitem{brossardUnscented2018}
M.~Brossard, S.~Bonnabel, and A.~Barrau, ``Unscented {{Kalman Filter}} on {{Lie
  Groups}} for {{Visual Inertial Odometry}},'' in \emph{{IEEE IROS}}, 2018.

\bibitem{hartleyContactAided2018}
R.~Hartley, M.~G. Jadidi, J.~W. Grizzle, and R.~M. Eustice, ``Contact-{{Aided
  Invariant Extended Kalman Filtering}} for {{Legged Robot State
  Estimation}},'' in \emph{Robotics {{Science}} and {{Systems}}}, 2018.

\bibitem{wuVINS2017}
K.~Wu, C.~Guo, G.~Georgiou, and S.~Roumeliotis, ``{{VINS}} on {{Wheels}},'' in
  \emph{IEEE ICRA}, 2017, pp. 5155--5162.

\bibitem{zhengOdometry2018}
F.~Zheng, H.~Tang, and Y.-H. Liu, ``Odometry {{Vision Based Ground Vehicle
  Motion Estimation With SE}}(2)-{{Constrained SE}}(3) {{Poses}},'' \emph{IEEE
  Trans. on Cybernetics}, pp. 1--12, 2018.

\bibitem{ramanandanInertial2012}
A.~Ramanandan, A.~Chen, and J.~Farrell, ``Inertial {{Navigation Aiding}} by
  {{Stationary Upyears}},'' \emph{IEEE Trans. on Intelligent Transportation
  Systems}, vol.~13, no.~1, pp. 235--248, 2012.

\bibitem{skogZeroVelocity2010}
I.~Skog, P.~Handel, J.~O. Nilsson, and J.~Rantakokko, ``Zero-{{Velocity
  Detection}}—{{An Algorithm Evaluation}},'' \emph{IEEE Trans. on Biomedical
  Engineering}, vol.~57, no.~11, pp. 2657--2666, 2010.

\bibitem{dissanayakeAiding2001}
G.~Dissanayake, S.~Sukkarieh, E.~Nebot, and H.~Durrant-Whyte, ``The {{Aiding}}
  of a {{Low}}-cost {{Strapdown Inertial Measurement Unit Using Vehicle Model
  Constraints}} for {{Land Vehicle Applications}},'' \emph{IEEE T-ROA},
  vol.~17, no.~5, pp. 731--747, 2001.

\bibitem{solinInertial2018}
A.~Solin, S.~Cortes, E.~Rahtu, and J.~Kannala, ``Inertial {{Odometry}} on
  {{Handheld Smartphones}},'' in \emph{{{FUSION}}}, 2018.

\bibitem{atchuthanOdometry2018}
D.~Atchuthan, A.~Santamaria-Navarro, N.~Mansard, O.~Stasse, and J.~Solà,
  ``Odometry {{Based}} on {{Auto}}-{{Calibrating Inertial Measurement Unit
  Attached}} to the {{Feet}},'' in \emph{ECC}, 2018.

\bibitem{yanRIDI2018}
H.~Yan, Q.~Shan, and Y.~Furukawa, ``{{RIDI}}: {{Robust IMU Double
  Integration}},'' in \emph{ECCV}, 2018.

\bibitem{wagstaffLSTMBased2018}
B.~Wagstaff and J.~Kelly, ``{{LSTM}}-{{Based Zero}}-{{Velocity Detection}} for
  {{Robust Inertial Navigation}},'' in \emph{IPIN}, 2018.

\bibitem{cortesDeep2018}
S.~Cortes, A.~Solin, and J.~Kannala, ``Deep {{Learning Based Speed Estimation}}
  for {{Constraining Strapdown Inertial Navigation}} on {{Smartphones}},''
  \emph{IEEE MLSP workshop}, 2018.

\bibitem{chenIONet2018}
C.~Chen, X.~Lu, A.~Markham, and N.~Trigoni, ``{{IONet}}: {{Learning}} to
  {{Cure}} the {{Curse}} of {{Drift}} in {{Inertial Odometry}},'' in
  \emph{{AAAI}}, 2018.

\bibitem{chenOxIOD2018}
C.~Chen, P.~Zhao, C.~X. Lu, W.~Wang, A.~Markham, and N.~Trigoni, ``{{OxIOD}}:
  {{The Dataset}} for {{Deep Inertial Odometry}},'' 2018.

\bibitem{brossardlearning2019}
M.~Brossard and S.~Bonnabel, ``{L}earning {W}heel {O}dometry and {IMU} {E}rrors
  for {L}ocalization,'' in \emph{IEEE ICRA}, 2019.

\bibitem{haarnojaBackprop2016}
T.~Haarnoja, A.~Ajay, S.~Levine, and P.~Abbeel, ``Backprop {{KF}}: {{Learning
  Discriminative Deterministic State Estimators}},'' in \emph{NIPS}, 2016.

\bibitem{barfootAssociating2014}
T.~Barfoot and P.~Furgale, ``Associating {{Uncertainty With
  Three}}-{{Dimensional Poses}} for {{Use}} in {{Estimation Problems}},''
  \emph{IEEE T-RO}, vol.~30, no.~3, pp. 679--693, 2014.

\bibitem{kingmaAdam2014}
D.~P. Kingma and J.~Ba, ``Adam: {{A Method}} for {{Stochastic Optimization}},''
  in \emph{ICLR}, 2014.

\end{thebibliography}
	
\end{document}